\pdfoutput=1

\documentclass[11pt]{article}

\usepackage[final]{acl}
\usepackage{todonotes}
\usepackage{tabularx}
\usepackage{colortbl}
\usepackage{times}
\usepackage{latexsym}
\usepackage{multicol}

\usepackage[T1]{fontenc}

\newcommand{\hlrow}{\rowcolor{gray!30!}}
\newcommand{\modelvshuman}{\hlrow \multicolumn{2}{l}{\textbf{Builder (model vs. original human dialogue):}}}

\newcommand{\fig}[3][0.9]{
\begin{figure}
\centering
\includegraphics[width=#1\columnwidth]{images/#2}
        \caption{#3}~\label{fig:#2}
\end{figure}}

\newenvironment{dialog}{\tabularx{\columnwidth}{>{\bfseries}r X}}{\endtabularx
    }

\usepackage[utf8]{inputenc}

\newcommand{\ACRO}[1]{\textsc{#1}}
\newcommand{\AMR}{\ACRO{amr}}

\newcommand{\MDC}{\ACRO{mdc}}

\usepackage{enumitem}
\setlist[description]{noitemsep, nolistsep}

\usepackage{microtype}

\usepackage{inconsolata}

\usepackage{graphicx}

\title{Referential ambiguity and clarification requests: comparing human and LLM behaviour}

  \author{Chris Madge, 
   Matthew Purver \and Massimo Poesio\\
  Queen Mary University of London \\
  \texttt{\{c.j.madge,m.purver,m.poesio\}@qmul.ac.uk} \\}

\begin{document}
\maketitle
\begin{abstract}
In this work we examine LLMs' ability to ask clarification questions in task-oriented dialogues that follow the asynchronous instruction-giver/instruction-follower format. 
We present a new corpus that combines two existing annotations of the Minecraft Dialogue Corpus --- one for reference and ambiguity in reference, and one for SDRT including clarifications --- into a single common format providing the necessary information to experiment with clarifications and their relation to ambiguity. 
With this corpus we compare LLM actions with original human-generated clarification questions, examining how both humans and LLMs act in the case of  ambiguity. 
We find that there is only a weak link between ambiguity and humans producing clarification questions in these dialogues, and low correlation between humans and LLMs. 
Humans hardly ever produce clarification questions for referential ambiguity, but  often do so for task-based uncertainty. 
Conversely,  LLMs produce more clarification questions for referential ambiguity, but less so for task uncertainty. 
We question if LLMs' ability to ask clarification questions is predicated on their recent ability to simulate reasoning, and test this with different reasoning approaches, finding that reasoning does appear to increase question frequency and relevancy.

\end{abstract}

\section{Introduction}

Large Language Models (LLM) are much maligned for their tendency to act presumptively, ``hallucinating'' in the absence of knowledge. Until the recent advent of reasoning, models struggled asking clarification questions \cite{deng2023prompting,li2022asking}.  Prior works have tested how LLMs respond to uncertainty, and proposed benchmarks \cite{zhang2024clamber}. However, this remains a challenge.

This is perhaps further complicated as clarification is a conversational strategy applied sparingly by humans \cite{purver2003means,Rodriguez04,Rieser05}.
Certain situations promote greater clarification question usage; 
for example, situations in which information is asymmetric, and which concerns a task requiring information seeking. This happens to be a popular paradigm for tasks created with the objective of soliciting dialogue (sometimes referred to as instruction giver/instruction follower) and also, in recent years, for studying clarification questions \cite{chi2020just,madureira2023you,testoni2024asking,shen2023learning}. 
In this work we look at the Minecraft Dialogue Corpus \cite{narayan2019collaborative}, a task orientated, grounded corpus that follows this paradigm.  We select this corpus as it has benefited from multiple separate annotation efforts \cite{thompson2024discourse,madge2025mdc} that extend its already richly structured offering, with useful supplemental information that can inform the experiments undertaken in this work.

We focus on one particular area of linguistic uncertainty, referential ambiguity.  There has been a long standing interest in reference, with ambiguity featuring as an interest in the first popular corpora \cite{pradhan2012conll}.   Our first contribution of this work is combining prior annotation efforts providing annotations for clarification questions \cite{thompson2024discourse} and reference \cite{madge2025mdc}
into a single aligned corpus in the MMAX format (described in Section \ref{sec:combined}).  

This annotation supports our next contribution, a comparison of how LLMs and humans resolve uncertainty.  We ask, \textit{``does referential ambiguity really trigger clarification requests from humans, and is this different for LLMs?''}.
We look at both the annotated instances of linguistic ambiguity, and the original clarification questions, as posed by human interlocutors for correlation. We test both against different LLM based approaches. 

One proposed approach to improving clarification questions with LLMs is through adding further reasoning capabilities with variations on the Chain of Thought approach \cite{deng2023prompting}. Despite extensive testing in epistemic, aleatoric, linguistic uncertainty \cite{ortega2023linguistic} and proposed benchmarks \cite{zhang2024clamber}, it remains somewhat unclear how effective LLMs are in identifying uncertainty and even more so how LLMs may consistently generate the relevant questions to address it.  

For our second research question, we ask if the ability to ask a useful clarification question, or indeed judge when to ask a question, is based on a model's ability to simulate reasoning.  We test this hypothesis with an experiment, comparing models that were trained to include reasoning and prompt engineering strategies for inducing reasoning at test time against ordinary models/methods.

For our final contribution, we look further into human reasoning and its constituent parts, with a discussion on how these may affect LLMs seemingly emergent ability to ask some clarification questions, and liken this to human reasoning.

\section{Related Work}

\subsection{Clarification Questions}

There has been a great interest in clarification questions in the literature  on dialogue systems going back at least twenty years \cite{purver2003means,schlangen2004causes,gabsdil2003clarification}. More recently, there has been extensive work in modern Natural Language Processing modelling clarification question generation or indeed when to ask them \cite{majumder2021ask,aliannejadi2019asking,kiseleva2022interactive}.  This section will primarily focus on prior works that target clarification question with overlap to our specific goals (i.e. task orientated dialogue with situated and/or embodied agents).
Previous works have gathered or annotated datasets in situated dialogue with clarification questions. For example, \cite{gervits2021should} gather a corpus (HuRDL - Human-Robot Dialogue Learning) and annotate clarification questions in a dialogue gathered from human participants in a robot situated tool gathering task.  \cite{gella2022dialog} annotate dialogue acts in the TEACh (Task-driven Embodied Agents that Chat) dataset \cite{padmakumar2022teach}; the product of a task that has human participants collaborating to perform household tasks in a virtual house environment.

A particularly popular task/dataset for this is CoDraw \cite{kim2017codraw}.  The CoDraw task \cite{kim2017codraw} is similar to the previously discussed Minecraft task, in that an instruction giver communicates with an instruction follower to collaboratively reach a goal.  As opposed to constructing a 3D voxel based structure, they recreate a scene formed of clipart images. \cite{madureira2023instruction} annotate this dataset with clarification questions. 

Previous works have also compared when humans and models would ask clarification questions
\citep{testoni2024asking} use the aforementioned CoDraw dataset to investigate this the relationship between model uncertainty, and human clarification questions based on task properties (e.g. size, orientation, position etc.). The presence of a clarification question is used as the measure of measure uncertainty, and they use logistic regression to see if they can predict this.

The Minecraft Dialogue Corpus \cite{narayan2019collaborative} used in this work is different, in that rather than referencing direct objects, continuously changing abstract shapes are created and manipulated during the dialogue. We expand on this further in the following section. There has been other work using Minecraft-like environments as a test-bed for the study of clarification questions in dialogue. However, this was prior to LLMs and looked at clarification question production as a task of ranking available clarification questions, rather than their generation \cite{kiseleva2022interactive}

Several works have investigated the use of LLMs for clarification question generation, with methods including: 
fine tuning on question data \cite{andukuri2024star};
uncertainty estimation over multiple samples \cite{pang2024empowering,zhang2023clarify} and
multi turn prompting strategies \cite{kuhn2022clam,li2023eliciting}. To our knowledge, none of these are primarily concerned with reference or situated dialogue settings.  There is however evidence to suggest LLMs can successfully resolve reference with performance similar to, or in some cases superior to, reference specific models \cite{hicke2024lions,le2023largelanguagemodelsrobust}.


%
\subsection{MDC and extensions}
The Minecraft Dialogue Corpus \cite{narayan2019collaborative} is a collection of conversations among human participants performing the Minecraft Collaborative Building Task.
This follows the typical instruction giver, instruction follower paradigm, where \textit{the Architect}, who has full observability over the target environment but is unable to act, instructs \textit{the Builder}, to manipulate the environment to meet that target structure.  The world is a 
 3D voxel based $11 \times 9 \times 11$ Minecraft like world, originally provided by project Malmo \cite{johnson2016malmo}.
 This results in a 509 multi turn situated dialogues with rich linguistic phenomena including reference and clarification.

Various annotation efforts have extended {\MDC}, including variations of {\AMR} \citep{bonn2020spatial,bonial2021builder}, reference \citep{madge2025mdc} and Segmented Discourse Representation Theory  \citep[\ACRO{sdrt},][]{thompson2024discourse}.  We focus on the latter two as they are directly used in this work.

SDRT provides a macrostructure of interconnected logical discourse forms, linking narrative arcs and discourse relations \citep[e.g.\ corrections, confirmations, acknowledgements etc., see][]{asher2003logics,lascarides2007segmented}. \citet{thompson2024discourse} exhaustively annotated {\MDC} with SDRT in their Minecraft Structured Dialogue Corpus (MSDC).

MDC-R \cite{madge2025mdc} annotates a subset of 100 dialogues from {\MDC} with expert annotated reference according to ARRAU \cite{poesio-et-al:ARRAU3:manual}. The dynamically changing environment and instruction based two-party dialogue gives rise to various types of reference, much of which, beyond the discourse, is linked directly to the objects in the virtual world.  This results in some interesting and challenging examples of ambiguity for a dialogue system to resolve.

\subsection{Reasoning in Large Language Models}

Chain of Thought \citep[COT,][]{wei2022chain} simulates reasoning at inference time by encouraging the model to think through the answer step by step. In implementation, this can take one of two common forms.  The model is either provided an example of thinking through a problem step by step as part of a one-shot/few-shot prompt, or a zero-shot approach that simply prefaces the prompt with something like ``Let's think this through, step by step...''. The core benefits of the COT approach are given to be: problem decomposition; some explainability/insight into how results are reached; logical problem solving/symbolic manipulation and ease of application to existing models. Previous work has observed an improvement in applying this method when addressing Minecraft orientated tasks \cite{madge2024llm}.
Following Chain of Thought, several models have been trained or aligned explicitly to follow this process (e.g. Gemini 2.5; Version 3 of the Qwen model \cite{qwen2.5}; DeepSeek-R1 \cite{guo2025deepseek}).

\section{Adding MSDC information to MDC-R}\label{sec:combined}

In this section we will motivate and describe our effort to add MSDC information to  MDC-R to produce a new version of the corpus combining both types of annotation.


We identify two types of utterances or phrases may provoke clarification requests.  Firstly, utterances that have been annotated as the subject of a clarification or confirmation request in the dialogue  (typically related to task orientated uncertainty), and secondly instances of referential ambiguity.

 To support our experimentation, and investigation of any possible relationship between the two, we present a corpus that merges two existing corpora that identifies these.  This is a combination of the previously discussed MDC-R \cite{madge2025mdc}, and MSDC corpora \cite{thompson2024discourse}, permitting convenient examination of reference and more specifically types of referential ambiguity aligned with clarification questions.    We add a new MMAX\footnote{\url{https://mmax2.net/}} layer, referred to as SDRT. Each MMAX markable in this layer represents an Elementary Discourse Unit and the relations between those markables are represented by a \textit{to} attribute on each markable, with the related markables unique ID. This combined format allows parsing and examination of adjacent reference (shown in Figure \ref{fig:lingambig}) and clarification (shown in Figure \ref{fig:cq}).

 \fig[1]{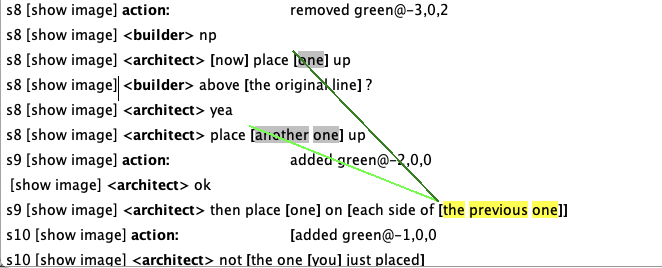}{Referential Ambiguity Annotation Example in MMAX}
 
 \fig[1]{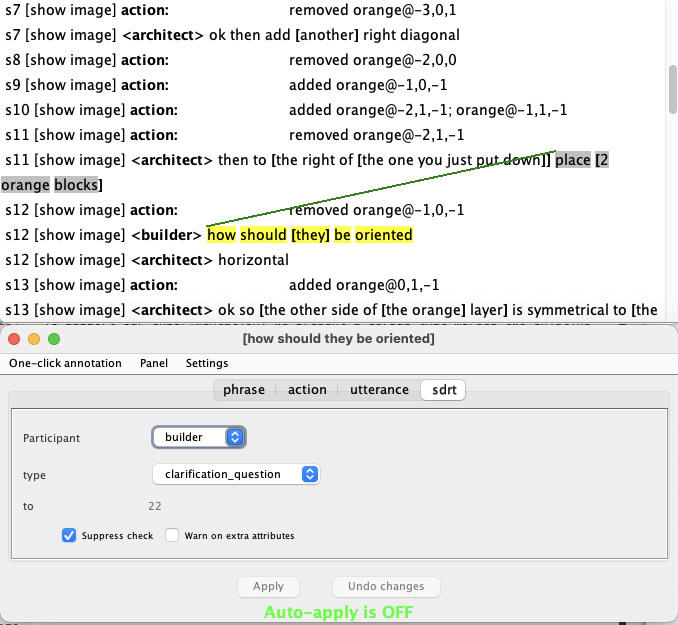}{Clarification Question Annotation Example in MMAX}

 Reference annotation has many parameters, with each relationship holding many attributes.  Whilst the SDRT annotations exhaustively cover all of the original MDC dataset \cite{narayan2019collaborative}, MDC-R covers a subset of 100 dialogues (some more detailed descriptive statistics taken from MDC-R \cite{madge2025mdc} are given in Table \ref{tab:stats}).  As such, our corpus will be limited to the same 100 dialogues.

 \begin{table}[h!]
    \centering
    \begin{tabular}{rc|rc}
         Statistic &  Count &  Statistic &  Count  \\
         \hline
         Documents &  101 &
         Tokens & 29,174 \\ 
         Utterances & 3,343 & 
         Actions & 5,793 \\ 
         Markables &  7,600 & 
         Discourse old &  1960 \\ 
         Bridging & 1,053 & 
         Discourse Deixis &  500 \\ 
         Plural &  24 & 
         Ambiguous &  149 \\ 
    \end{tabular} 
    \caption{MDC-R Corpus Statistics \cite{madge2025mdc}} \label{tab:stats}
\end{table}
%
 

Table \ref{tab:frequency} shows the frequency of the different instances of subjects that could motivate a question as a percentage of utterances in the selected subset of the corpus.  

\begin{table}[h!]
    \centering
    \begin{tabular}{r|c|c}
         Type & Instances & \% \\
         \hline
confirmation question	& 218 & 3.7\% \\ 
clarification question	& 182 & 3.1\% \\ 
discourse	& 24 &  0.4\% \\ 
space	& 16 & 0.3\% \\ 
    \end{tabular}
    \caption{Frequency of instances as a percentage of utterances}
    \label{tab:frequency}
\end{table}

77\% of dialogues contain a clarification question and 75\% a confirmation question.

The SDRT annotations of the complete corpus found in MSDC had 999 confirmation questions and 960 clarification questions over 547 dialogues.  We can see from the relative quantity of questions, that the selected dialogues do appear to be representative of the corpus as a whole, with respect to question quantity.

To give some overview, the most common phrases for discourse deixis are: \textit{``that''} (8); \textit{``this''} (7) and \textit{``it''} (3).  Spatial deixis has 3 instances of ``this'', but referents while still ambiguous, tend do be more literal (e.g. \textit{``the red end''}).  We expand further on this in a discussion of reoccurring patterns in Section \ref{sec:lingresults}.

The corpus is available at \url{https://github.com/arciduca-project/MDC-R/tree/sdrt}.

\section{Methodology}

To test how LLMs perform clarification questions, we first discover points in the conversation that may require clarification, then we sample from various LLMs with different approaches, feeding the context of the conversation up to the appropriate point. 
Examples of our prompts are given in the Appendix (see Section \ref{sec:prompts}). A system prompt describes the nature of environment and it's constraints. This differs slightly between architects and builders, in that architects can see the target structure, and the builder's system prompt specifies the required JSON response format necessary to encode their resultant actions. For the chain-of-thought treatments, we supplement these system prompts as follows.  The zero-shot chain-of-thought approach simply adds, ``Think step by step'' to the system prompt.  The one-shot approach follows the system prompt with an example exchange that incorporates thinking.
These approaches are deliberately selected to compare reasoning based models (e.g. Llama3, DeepSeek-R1 \cite{guo2025deepseek}) and sampling methods (COT, zero-shot and one-shot) vs. non-reasoning.  We compare the approach taken by LLMs to the approach originally taken by humans.

There are three characteristics of the instances that we identify to test against.
Firstly, the subject of any clarification and confirmation question as originally annotated in SDRT.  Second and thirdly, linguistic ambiguities.
We select any referent that has two or more antecedents, whether they be discourse deixis (part of the discourse), or spatial deixis (in reference to objects in the environment.  That is to say, in the MDC-R MMAX format, for discourse deixis the phrase in question, would have the following attributes specified \texttt{segment\_phrase\_antecedent\_2}, and for spatial deixis, \texttt{object2} specified.

We also check for correlation between clarification questions, and the instances of linguistic ambiguity, as permitted by using our new merged corpus (described in Section \ref{sec:combined}), 

We measure the tendency of different approaches to ask a question and use a single human coder to examine specific instances of ambiguity and their responses to attempt to quantify the number of relevant questions asked when applying each method.   Prior to quantifying responses to instances we conduct a thorough example driven investigation.

\section{Results}

In the first section we look at how LLMs and humans respond to instances of referential linguistic ambiguity, including whether humans pose clarification questions when they encounter linguistic ambiguity. In the second section we look at the original clarification questions as posed by humans, to see how LLMs respond. Finally, we perform a quantitative evaluation, counting the tendency of different approaches to ask questions, and a count of question relevancy under one specific condition.


\subsection{Linguistic Ambiguities}\label{sec:lingresults}

\paragraph{Do humans ask clarification questions when there is ambiguity?}

In many cases, where linguistic ambiguity exists, it appears deliberately underspecified in the interest of brevity, and where expanding more literally or verbosely would not have any further positive effect on task completion. Here, we see two common patterns emerge.

The first common pattern is in the communication of approval.  The exact target of approval is unspecified, and arguably redundant if correct, as no further intervention is required. This appears to be used as a signal to indicate objective completion and generally precede new instruction. Examples include: \textit{``oh that is magnificent builder''}; \textit{``... that is great''}; \textit{``...that's perfect''}; \textit{``...let 's see what we can do about that''}; \textit{``yeah, that's fine''}; \textit{``good job''}, \textit{``yes like that''}. 

These seem closely related to the second commonly appearing pattern in sources of linguistic ambiguity,  which occur in clarification questions themselves and appear to be used as a polite conversational device to invite architect interruption to general correction and indicate the builder believes the prior instruction to be addressed.  Examples include: ``\textit{like this?}''; ``\textit{is this good?}''; ``\textit{this right?}'', where ``this'' and ``that'' do not have clear referrents.  Again, exact specification of what ``this'' refers to is somewhat redundant, as any corrective instruction may follow.
Examining the context an interpreting more explicitly, these would appear to be more semantically equivalent to: ``I believe I've completed all instructions, I'm ready for the next instruction or correction''.

The first finding of this paper is that linguistic ambiguities are  rarely  the target of clarification questions directly. We only discovered a single instance in which a discourse deixis referential ambiguity is in the vicinity of the subject of a latter clarification question.  In the case of linguistic ambiguity, it seems humans prefer to act presumptively, in this setting.

\paragraph{Do LLMs ask clarification questions when there is ambiguity?}

 Unresolved linguistic ambiguities, left unchecked, do propagate, resulting in the necessity for clarification or confirmation later in the conversation. Whilst human participants do not address these immediately, LLMs sometimes do. We show two examples here. In the following example, ``the previous one'' has an ambiguous referent in the context of the conversation and current world state.  This is not immediately resolved at the time by the human builder, who acts presumptively, and takes action.  Having taken the incorrect action, and then been required to perform a revision, the builder then uses a confirmation question, \textit{``like that?''}. In contrast, the LLM however, does ask a question to directly disambiguate, \textit{``the previous one''} by offering the two possible interpretations (see Figure \ref{fig:propref}).
 

\begin{figure}[!ht]
\small
\begin{dialog}
Architect & then place one on each side of \underline{the previous one} \\
\modelvshuman \\
\hlrow Model  & Are the blocks to be placed on each side of the green block at [-2,0,0] or in relation to the vertical stacking process near the center? \\
\hlrow Human & \textit{takes action} \\
Architect & not the one you just placed \\
Builder & \textit{revisits with corrective action} \\
Builder & \underline{like that?} \\
\end{dialog}
\caption{Excerpt from B4-A19-C4-1522882189483 demonstrating propagation of referential ambiguity}
\label{fig:propref}
\end{figure}


Similarly, in Figure \ref{fig:immcq}, the human builder does not clarify the ambiguous statement, ``the same thing'', but proceeds to take action, to discover a lot of work must then be mirrored.  Conversely, an LLM asks, about which plane the blocks should be mirrored, to first discover the correct approach.

\begin{figure}[ht!]
\small
\begin{dialog}
Builder & \textit{builder takes action} \\
Architect & now \underline{the same thing} on the other side  \\
\modelvshuman \\
\hlrow Model & Does `the same thing on the other side' mean mirroring the blocks symmetrically with respect to the y-z plane at x=0? \\
\hlrow Human & \textit{builder takes further action} \\
Architect & that but on the other side  \\
Builder & \textit{builder carries out extensive revision} \\
Architect & there \\
\end{dialog}
\caption{Excerpt from B35-A34-C118-1524004322117 demonstrating direct and immediate clarification of linguistic ambiguity from LLM}
\label{fig:immcq}
\end{figure}

We show here in this section that whilst linguistic ambiguities rarely trigger direct human clarification, despite the fact they may translate to uneconomical task action, LLMs can and do sometimes respond with targeted clarification questions.

\subsection{Clarification/Confirmation Questions}

Having shown in the previous section, there is not a strong correlation between linguistic ambiguity and clarification questions, this section looks, at under what circumstances clarification questions are asked by human participants, and how this behaviour compares with how LLMs respond.  

A common repeated pattern by human builders is the use of questions such as, \textit{``like that?''}, as previously discussed, and others that are potentially more specific, such \textit{``is this right?''} and \textit{``how's that''}, with contextually evident reference to specific actions.  This is not a behaviour employed by the LLMs tested here.  As such, these clarification questions are unique to human participants.

Another clarification question approach exhibited uniquely by humans, is that of interrupting a set of instructions to proactively anticipate a possible pattern that they have observed to potentially reduce further instruction.  For example: \textit{``Are we repeating the structure?''}; \textit{``Are we repeating the pattern with orange and yellow?''}; \textit{``both sides or just this one?''}; \textit{``are we going to repeat what we did on the other side?''}; \textit{``same design? just blue?''}; \textit{``repeat here?''}.

Clarification questions typically occur not directly around linguistic ambiguity, but rather ambiguity in relation to the task objectives (e.g. properties).  In the following example, the architect specifies colour and quantity, but does not specify orientation, leading to similar, albeit more concise on part of the human, questions from human and LLM (Figure \ref{fig:orientcq}). Similarly, for individual block position in Figure \ref{fig:poscq}.

\begin{figure}[ht!]
\small
\begin{dialog}
Architect & then to the right of the one you just put down place 2 orange blocks  \\
\modelvshuman  \\
\hlrow Model &	What is the current orientation of the blocks that have been placed so far? Are they forming a diagonal line from north to south or from east to west? \\
\hlrow Human & how should they be oriented \\
\end{dialog}
\caption{Excerpt from B19-A36-C118-1524093668453 showing the subject of a clarification request  in relation to orientation uncertainty}
\label{fig:orientcq}
\end{figure}

\begin{figure}[ht!]
\small
\begin{dialog}
Architect & great now one orange below that one  \\
\modelvshuman \\
\hlrow Model &	Is it correct to assume 'now one orange below that one' means in the same diagonal direction as 'now one red block down along the diagonal \\
\hlrow Human & diagonally or nah?  \\
\end{dialog}
\caption{Excerpt from B16-A29-C144-1524067263550 showing the subject of a clarification request in relation to individual position uncertainty}
\label{fig:poscq}
\end{figure}

There are several examples in relation to collective, individual, absolute and relative property. LLMs seem to also approach these well.  Another common example is colour (shown in Figure \ref{fig:colorcq}). Again, the LLM reaches a fair clarification question.

\begin{figure}[ht!]
\small
\begin{dialog}
Architect & on the yellow block touching the orange , add two blocks to either side of it . making a t.  \\
\modelvshuman \\
\hlrow Model & What color are the new blocks to form the T shape and do you want me to remove or replace any existing blocks at [-4,1,0], [-3,0,0], [-2,1,0]? \\
\hlrow Human & what color blocks ?  \\
\end{dialog}
\caption{Excerpt from B37-A23-C102-1523629957597 showing the subject of a clarification request in relation to the colour property}
\label{fig:colorcq}
\end{figure}


\subsection{Quantitative Analysis}

In this section we aim to quantify the differences between approaches and how they respond to situations that may benefit from clarification.

Table \ref{tab:questions} looks at the number of questions asked in each reply.  In this first case, we do not assess question relevance, but simply the tendency to reply with a question. We see a lower tendency to ask a question in the previous generation models that did not feature reasoning.  However, later models that do feature reasoning, or previous generation models supplemented with prompts that attempt to simulate reasoning, perform comparably with a greater tendency to question. 

\begin{table}[!ht]
    \centering
    \begin{tabular}{r|c}
         Approach & Questions Asked \\
         \hline
         llama2:13b & 276 \\
         llama2:13b-COTZERO & 303 \\
         llama2:13b-COTONE & 327 \\
         deepseek-r1:8b & 354 \\
         llama3.2:3b & 278 \\
         llama3.3:70b & 383 \\
    \end{tabular}
    \caption{Number of questions asked in different approaches}
    \label{tab:questions}
\end{table}

To provide some notion of relevancy, Table \ref{tab:rquestions} looks specifically at the number of questions asked that target the instance of expert annotated ambiguity compared with the actions of the original human participant in the conversation.  These are counted solely for instances of spatial deixis, as these are very literal and therefore the easiest to objectively assess. For example, in \textit{``now a tower of five oranges on top of \underline{the red end}''}, Llama3.3:70b's response of \textit{``Which end of the red blocks is considered the 'red end', the one at coordinates [3, 0, 1] or [0, 0, 1]?''}, is considered to target the ambiguity.

We see that the previous generation models, not trained for reasoning, but with Chain of Thought prompting, perform comparably to modern reasoning orientated models, in this regard.  Human participants did not choose to disambiguate these phrases direct at the time, but rather acted presumptively. This does not measure the final utility of asking the question, or any impact it may have on the conversation.

\begin{table}[!ht]
    \centering
    \begin{tabular}{r|c}
         Approach & Spatial Deixis\\
         \hline
         Human participant & 0 \\
         llama2:13b & 0 \\
         llama2:13b-COTZERO & 1 \\
         llama2:13b-COTONE & 3 \\
         deepseek-r1:8b & 3 \\
         llama3.2:3b & 2 \\
         llama3.3:70b & 5 \\
    \end{tabular}
    \caption{Number of relevant spatial deixis questions asked by approach}
    \label{tab:rquestions}
\end{table}

Another limitation to our experiment.  It's challenging to communicate a sense of perspective to the LLMs.  As a consequence, some instructions do not make sense, e.g.: \textit{``A:  can you come to the side of the structure so you have a side view''}; \textit{``B: left or right''}; \textit{``B: forward to my right or in front of me?''}; \textit{``B: this perspective ?''}.

\section{Reasoning and Clarification}

Our experiments would appear to show that the recent advent of reasoning in models has the emergent benefit of allowing models to ask clarification questions.  In this section, following our hypothesis that clarification is dependent on an ability to reason,
we look at clarification questions in relation to reasoning through the lens of human psychology, and where available, assessments of LLM abilities to perform the required components of related human reasoning abilities

Examining the role of reasoning in clarification, knowing when to ask a question, requires reflection on the gaps in ones own knowledge, or a higher order of thought, referred to in human psychology as \textbf{Metacognition} \cite{flavell1979metacognition}.  LLMs originally lacked any awareness of gaps in their knowledge, acting presumptively, leading to ``hallucination'' where, albeit often grammatically valid, a model's output would be factually incorrect or possibly nonsensical. This is most commonly due to a lack of knowledge \cite{zhang2023siren}, or perhaps the inability to reason when knowledge is absent. Metacognition is the ability to reflect on held knowledge by self questioning. The Metacognitive capabilities of LLMs have been explored in previous works in relation to reasoning \cite{didolkar2024metacognitive}. Metacognitive prompting that, self questions to enhance reasoning \cite{wang2023metacognitive} has been explored in LLMs, as has self-questioning with the goal of reducing hallucination \cite{dhuliawala2023cove}. A model knowing whether it has the applicable knowledge or skill to proceed, or whether to direct its process to clarification, could be seen as metacognitive regulation.

When an agent is working in collaboration with other interlocutors, clarification may be dependent on discovering other parties knowledge, abilities or attitudes and approach to a task.
This crucial component, in psychology, is an aspect of reasoning known as \textbf{Theory of Mind} \cite{premack1978does}, which relates to reasoning about other participants belief states. There is some evidence to suggest, as a consequence of simulated reasoning ability, LLMs may now be able to simulate this also \cite{kosinski2023theory}. This has been explored with LLMs in Mindcraft, which is a collaborative task in which the players have separate skills and must negotiate to reach a common goal \cite{bara2021mindcraft}.  This is particularly important for LLMs in referential communication \cite{sidera2018theory}



%
%
%
%




\section{Conclusion}

To conclude, we find that humans do not commonly ask a clarification question when language is ambiguous.  In the majority of these cases the instances of linguistic ambiguity appear deliberate in the interest of conversational efficiency. The resultant utility of a clarification question is not examined in this work, and may be the subject of a future work. Consequently, there may be little utility to asking a question in many of those situations. We did however identify one situation in which that uncertainty did propagate to create future issue. Regardless of conversational efficiency, we do find that LLMs, particularly reasoning orientated ones, are capable of asking relevant clarification questions under those circumstances.

On the topic of the originally posed human clarification questions, we found they largely followed a specific pattern and strategy not adopted by the LLMs.  That was, to perform actions then use a clarification question to verify they were correct. These instances of human clarification questions largely relate to task based ambiguity rather than linguistic ambiguity.    

Across all instances, we find a greater tendency of reasoning orientated approaches to pose clarification questions and find that this can be somewhat matched at test time, with methods such as COT.

\section*{Acknowledgements}

This research was funded by ARCIDUCA, EPSRC EP/W001632/1

\bibliography{custom}

\appendix

\section{Appendix}

\subsection{System Prompts}\label{sec:prompts}

\paragraph{Architect}

You are an agent in a voxel world, where the most northernly point is 0,0,-5; the most westerly point -5,0,0; the most eastern point is 5,0,0; the most southern 0,0,5 and the y-axis is up and down, with y=0 being the minimum. Your task is deliver instructions and respond to questions where necessary, to achieve a target world state []. The current world state is []. They should be easy to understand instructions, Do not ask for specific coordinates and do not accept them. Omit any pleasantries. You may ask questions

\paragraph{Builder}

You are an agent in a voxel world, using the following coordinate system, where the center on the ground is [0,0,0], [0,9,0] is the highest point in the air, the most northernly point is [0,0,-5]; the most westerly point [-5,0,0]; the most eastern point is [5,0,0]; the most southern [0,0,5]. Describe the coordinates of the blocks their colours (must be one of: blue, yellow, green, orange, purple, red) and whether the action is to add or remove them. You MUST include a question if the instruction is potentially unclear or ambiguous and you MUST follow the final JSON format: {"add": [[x,y,z,"color"], ...], "remove": [[x,y,z,"color"], ...], "confidence": 0.0, "question": "..."}.  The current world state is []. Give your final response in JSON only, without additional dialog. You may ask questions

\subsection{COTONE}
\begin{description}
\item[user]
\begin{itemize}
    \item builder: hi, let's start
    \item architect: place three blue blocks in a row in the center. Place a red block on top of the blue block
\end{itemize}
\item[assistant] 
\begin{itemize}
    \item architect: \textbf{Thinking:} let's think. we know three blue blocks need to go in the center.  The center is [0,0,0], so to place three blue blocks would be [-1,0,0,"blue"],[0,0,0,"blue"],[1,0,0,"blue"]. Now we need to place a red block on top of the blue block. Wait, they didn't tell us which blue block, there are three.  We must ask a question also. ``Which of the three blue blocks would you like the red block placed on?''. We need to formulate this response in the given JSON format.
    \item \textbf{Response:} builder: \{"add": [[-1, 0, 0, "blue"], [0, 0, 0, "blue"], [1, 0, 0, "blue"]], "confidence": 0.5, "question": "Which of the three blue blocks would you like the red block placed on?"\} 
\end{itemize}
\end{description} 


\end{document}